\documentclass[11pt,a4paper]{article}
\usepackage{authblk}
\usepackage[nohyperref]{acl2019}
\usepackage{times}
\usepackage{latexsym}
\usepackage{amsmath}
\usepackage{cleveref}
\usepackage{graphicx}
\usepackage{booktabs}
\usepackage{multirow}
\usepackage{amsmath}
\usepackage{enumitem}
\usepackage{color}
\usepackage{float}
\usepackage{xcolor,colortbl}
\usepackage{url}

\aclfinalcopy 

\newcommand\blfootnote[1]{%
  \begingroup
  \renewcommand\thefootnote{}\footnote{#1}%
  \addtocounter{footnote}{-1}%
  \endgroup
}

\title{Out-of-Vocabulary Embedding Imputation with Grounded Language Information by Graph Convolutional Networks}

\author[1]{Ziyi Yang}
\author[2]{Chenguang Zhu}
\author[3]{Vin Sachidananda}
\author[1, 4]{Eric Darve}
\affil[1]{Department of Mechanical Engineering, Stanford University}
\affil[2]{Microsoft Speech and Dialogue Research Group}
\affil[3]{Department of Electrical Engineering, Stanford University}
\affil[4]{Institute for Computational and Mathematical Engineering, Stanford University}
\affil[ ]{\texttt {\{ziyi.yang, vsachi, darve\}@stanford.edu}, \texttt {chezhu@microsoft.com}}

\date{}

\begin{document}
\maketitle
\begin{abstract}
   Due to the ubiquitous use of embeddings as input representations for a wide range of natural language tasks, imputation of embeddings for rare and unseen words is a critical problem in language processing. Embedding imputation involves learning representations for rare or unseen words during the training of an embedding model, often in a post-hoc manner. In this paper, we propose an approach for embedding imputation which uses grounded information in the form of a knowledge graph. This is in contrast to existing approaches which typically make use of vector space properties or subword information. We propose an online method to construct a graph from grounded information and design an algorithm to map from the resulting graphical structure to the space of the pre-trained embeddings. Finally, we evaluate our approach on a range of rare and unseen word tasks across various domains and show that our model can learn better representations. For example, on the Card-660 task our method improves Pearson's and Spearman's correlation coefficients upon the state-of-the-art by 11\% and 17.8\% respectively using GloVe embeddings. \blfootnote{Published as a conference paper at ACL 2019}
  
\end{abstract}

\section{Introduction}
Word embeddings \citep{Mikolov:2013, Pennington14glove:global} are used pervasively in deep learning for natural language processing. However, due to fixed vocabulary constraints in existing approaches to training word embeddings, it is difficult to learn representations for words which are rare or unseen during training. This is commonly referred to as the out-of-vocabulary (OOV) word problem. In the original embedding implementations, a special OOV token is typically reserved for such words. However, this rudimentary approach often detriments the performance of downstream tasks which contain numerous rare or unseen words. Recent works have proposed subword approaches \citep{Zhao18, SennrichHB15}, which construct embeddings through the composition of characters or sentence pieces for OOV words. Vector space properties are also utilized to learn embeddings with small amounts of data \citep{BahdanauBJGVB17,HerbelotB17}. In this paper, we propose a novel approach, knowledge-graph-to-vector (KG2Vec), for the OOV word problem. KG2Vec makes use of the grounded language information in the form of a knowledge graph. Grounded information has been extensively used in various NLP tasks to represent real-world knowledge \citep{Niles03linkinglexicons, Gruber:1993:TAP:173743.173747, Guarino:1998:FOI:521669, bruijn, Paulheim17} . In particular, early question answering systems used expert-crafted ontologies in order to endow these systems with common knowledge \citep{Harabagiu2005EmployingTQ, Xu2016HybridQA}. Additionally, lexical-semantic ontologies, such as WordNet, have been used to provide semantic relations between words in a wide variety of language processing and inference tasks \citep{Morris:1991:LCC:971738.971740, OVCHINNIKOVA10.84}.

Grounded language information has been observed to augment model performance on a wide variety of natural language processing and understanding tasks \citep{GraphDialogue, Quac}. In these settings, a model is able to provide better generalization by using relational information from a knowledge graph or knowledge base in addition to the standard set of training examples. Additionally, outputs from models with grounded approaches have been observed to be more factually consistent and logically sound \citep{BordesCW14} compared with outputs from models without grounding information. 

By foregoing the usage of vector space or subword information, KG2Vec is able to capture semantic meanings of words directly from the graphical structure in grounded knowledge using recent advances in network representation learning. Furthermore, KG2Vec leverages the most updated information from comprehensive knowledge bases (Wikipedia \& Wiktionary). Therefore, KG2Vec can be applied to training embeddings of newly emerging OOV words. 

In summary, our contributions are three-fold:

\begin{enumerate}[noitemsep,topsep=0pt]
\item An approach to constructing graphical representations of entities in a knowledge base in an unsupervised manner.
\item Methods for mapping entities from a graphical representation to the space in which a pre-trained embedding lies. 
\item Experimentation on rare and unseen word datasets and a new state-of-art performance on Card-660 dataset.
\end{enumerate}

\section{Related Work}

\subsection{Graph Neural Networks}
Graph neural networks (GNN) are an emerging deep learning approach for representation learning of graphical data \cite{xu2018powerful, kipf2016semi}. GNNs can learn a representation vector $h_v$ for each node in the network by leveraging the graphical structure and node features $f_v$. Node embeddings are generated by recursively aggregating each node's neighborhood information and features. At the $t$-th iteration, the information aggregation is defined as:
\begin{equation}
    h_v^t = M^{t}(h_v^{t-1}, \{h_u^{t-1}\}_{u\in N(v)})
\end{equation}
where $h_v^t$ is the representation for $v$ at the $t$-th iteration, $M^{t}$ is an iteration-specific message aggregation function parametrized by a neural network and $N(v)$ is the set of neighbors of node $v$. One simple form of $M^t$ is mean neighborhood aggregation:
\begin{equation}
    h_v^t = \mbox{ReLU}(\sum_{u\in N(v)}\frac{W^th_u^{t-1}}{|N(v)|} + B^th_v^{t-1})
\end{equation}
where $W^t$ and $B^t$ are trainable matrices. Typically, $h_v^0$ is initialized as $f_v$. The final node representation is usually a function of $h_v^T$ from the last iteration $T$, such as an identity function or a transformation function \citep{ying2018hierarchical}.

\subsection{The OOV word problem}
The out-of-vocabulary (OOV) word problem has been present in word embedding models since their inception \citep{Mikolov:2013, Pennington14glove:global}. Due to space and training data constraints, words which are either infrequent or do not appear in the training corpus can lack representations at the time of inference. 

Numerous methods have been proposed to tackle the OOV word problem with a small amount of training data. Deep learning based approaches \cite{BahdanauBJGVB17} and vector-space based methods \cite{HerbelotB17} can improve the rare word representations on various semantic similarity tasks. One downside to these approaches is that they require small amounts of training data for words whose embeddings are being imputed and, as a result, can have difficulties representing words for which training samples do not exist.

Sub-word level representations have been studied in the context of the OOV word problem. \citet{PinterGE17} uses the RNN's hidden state of the last sub-word in a word to produce representations. \citet{Zhao18} proposes using character-level decomposition to produce embeddings for OOV words.

\section{Model}
 We propose the knowledge-graph-to-vector (KG2Vec) model for building OOV word representations from knowledge base information. KG2Vec starts with building a knowledge graph $\mathcal{K}$ with nodes consisting of pre-trained words and OOV words. It then utilizes a graph convolutional network (GNN) to map graph nodes to low-dimensional embeddings. The GNN is trained to minimize the Euclidean distance between the node embeddings to pre-trained word embeddings in the dictionary such as GloVe \citep{Pennington14glove:global} and ConceptNet Numberbatch \citep{speer2017conceptnet}. Finally, the GNN is used to generate embeddings for OOV words.

\subsection{Build the Knowledge Graph}
In a knowledge graph $\mathcal{K}$, each node $v$ represents a word $w_v$. The nodes (words) in the graph are chosen as follows. We count the frequency of occurrences for English words from the Wikipedia English dataset (with 3B tokens). The 2000 words with the highest frequencies of occurrence are skipped to diminish the effect of stop words. Among the words left, we choose the $|V^\prime|$ words with the highest frequencies of occurrence. All OOV words for which we would like to impute embeddings are also added to the graph as nodes.

For each node, we obtain its grounded information from two sources: (I) the words' summary, defined as the first paragraph of the Wikipedia page when this word is searched; (II) the word's definition in Wiktionary. We choose Wikipedia and Wiktionary over other knowledge bases because they are comprehensive, well-maintained and up-to-date. Here is an example of the grounded information for the word Brexit.
\begin{itemize}[noitemsep,topsep=0pt]
    \item Wikipedia page summary: \textit{Brexit, a portmanteau of ``British'' and ``exit'', is the impending withdrawal of the United Kingdom (UK) from the European Union (EU). It follows the referendum of 23 June 2016 when 51.9 per cent of voters chose to leave the EU...}
     \item Wiktionary definition: Brexit (Britain, politics) \textit{The withdrawal of the United Kingdom from the European Union.}
\end{itemize}
All the words in the Wikipedia summary and the Wiktionary definition form the grounded language information of this word $w_v$, defined as $D_v$. Specifically, $D_v$ is the concatenation of $w_v$'s Wikipedia summary and the Wiktionary definition. An undirected edge $e_{vu}$ exists between node $v$ and $u$ if the Jaccard coefficient $\frac{|D_v \cap D_u|}{|D_v \cup D_u|} > \eta$, where $\eta$ is a pre-defined threshold and chosen to be $0.5$ empirically in the experiments. The edge $e_{vu}$ is then assigned with a weight $s_{vu} = \frac{|D_v \cap D_u|}{|D_v \cup D_u|}$. We also compute a feature vector $f_u$ as the mean of pre-trained embeddings of words in $D_v$. Finally, the obtained knowledge graph $\mathcal{K} = (V, E)$ has a feature vector $f_v$ for each node $v\in V$.

\subsection{Graph Neural Network}
The nodes in the graph are mapped to low-dimensional embeddings via graph convolutional neural network (GCN) \citep{kipf2016semi}. It follows that, at the $t$-th neighborhood aggregation, the node embedding $h_v^t$ for node $v$ is modelled as:
\begin{equation}
    h_v^t = \text{ReLU}(W^t\sum_{u\in S(v)} \frac{s_{vu}h_u^{t-1}}{C} + b^t)
\end{equation}
where $S(v)=N(v)\cup\{v\}$, and the normalization constant $C = 1+\sum_{u\in N(v)}s_{vu}$. $W^t$ and $b^t$ are trainable parameters. The node embeddings are initialized as the feature vector $f_v$, i.e. $h_v^0 = f_v$. At the final iteration $T$, the generated node embeddings \{$h_v^T$\} are computed without the ReLU function. The loss function of the GNN model is the mean square error between the pre-trained word vectors and generated embedding $h_v^T$ for all words in the graph which are part of the model's vocabulary (e.g. GloVe). During inference, OOV words are assigned embeddings computed by the GNN.

\section{Experiments}
To evaluate our method's ability to impute embeddings, we conduct experiments on the following rare and unseen word similarity tasks.

\subsection{Card-660: Cambridge Rare Word Dataset}
Card-660 \cite{pilehvar2018} is a word-word similarity task with 660 example pairs involving uncommon words and provides a benchmark for rare word representation models. Card-660 has a inter-annotator agreement (IAA) measure of 0.90, which is significantly higher than previous datasets for rare word representation. Additionally, Card-660 contains examples from a disparate set of domains such as technology, popular culture and medicine.

\subsection{Stanford Rare Word (RW) Similarity}
The Stanford Rare Word (RW) Similarity Benchmark \citep{Luong13} is a word-word semantic similarity task including 2034 word pairs and tests the ability of representation learning methods to capture the semantics of infrequent words. Due to the probabilistic underpinnings of word embeddings, where distances between two words' representations are approximately proportional to their co-occurrence probability in a corpus, the authors found that rare words often have more noisy representations due to having fewer training samples. Although RW has a relatively low IAA measure of 0.41, the benchmark has been well-studied in previous literature.


\subsection{Results}
\definecolor{Gray}{gray}{0.95}
\newcolumntype{a}{>{\columncolor{Gray}}c}
\begin{table*}[h]
  \centering
  \setlength{\tabcolsep}{3.6pt}
 \scalebox{0.9}
 {
  \begin{tabular}{c l ra ra| c ra ra}
    \toprule
    \multicolumn{2}{c}{\multirow{2}{*}{\bf Model}} &  
    \multicolumn{2}{c}{\bf Missed words} & 
    \multicolumn{2}{c}{\bf Missed pairs} &  &
    \multicolumn{2}{c}{\bf Pearson $r$} &   
    \multicolumn{2}{c}{\bf Spearman $\rho$}  \\
    \cmidrule(lr){3-4}
    \cmidrule(lr){5-6}
    \cmidrule(lr){8-9}
    \cmidrule(lr){10-11}
    & & RW & \multicolumn{1}{c}{\textsc{Card}} & RW &  \multicolumn{1}{c}{\textsc{Card}} & & RW & \multicolumn{1}{c}{\textsc{Card}} & RW & \multicolumn{1}{c}{\textsc{Card}} \\
    \midrule
    \multicolumn{2}{l}{ConceptNet Numberbatch} & ~~5\% &  37\%  & 10\%& 53\%  & & 53.0 & 36.0 & 53.7 & 24.7  \\
    & + Mimick  &  0\%  & ~~0\%  & ~~0\%  & ~~0\% & & 56.0 & 34.2 & 57.6 &  35.6 \\
    & + Definition centroid   &  0\%  & 29\%  &  0\% & 43\%  & & 59.1 & 42.9 & 60.3 & 33.8 \\
    & + Definition LSTM  &  0\% & 25\%  & 0\%  & 39\% & & 58.6 & 41.8 & 59.4 & 31.7  \\
    & + SemLand  & 0\% & 29\% & 0\% & 43\% & & \bf \underline{60.5}  & 43.4  & \bf \underline{61.7}  & 34.3  \\
    & + BoS  & 0\% & ~~0\% & 0\% & 0\% & & 60.0  & 49.2  & \bf \underline{61.7}  & 47.6  \\
    \midrule
    & + Node features  & 0.02\% & ~~7\% & 0.04\% & 12\% & & 58.4  & 54.0  & 59.7  & 51.4  \\
    & + \textbf{KG2Vec}  & 0.02\% & ~~7\% & 0.04\% & 12\% & & 58.6  & \bf \underline{56.9}  & 60.1  & \bf \underline{54.3}  \\
    
    \midrule
    \multicolumn{2}{l}{GloVe Common Crawl} & 1\% & 29\% & 2\% & 44\% & & 44.0 & 33.0 & 45.1 & 27.3\\
    & + Mimick   & 0\%  & ~~0\%  & ~~0\%  & ~~0\% & & 44.7 & 23.9 & 45.6 & 29.5  \\
    & + Definition centroid  & ~~0\% & 21\% & 0\% & 35\% & & 43.5  & 35.2 & 45.1 &  31.7  \\
    & + Definition LSTM  & 0\% & 20\%  & ~~0\%  &  33\%  & & 24.0 & 23.0  & 22.9  & 19.6  \\
    & + SemLand  & 0\% & 21\%  & ~~0\%  & 35\% & &  44.3  & 39.5  & 45.8 & 33.8  \\
    & + BoS  & 0\% & ~~0\% & 0\% & ~~0\% & & \bf 44.9  & 31.5  & \bf 46.0  & 35.3  \\
    \midrule
    & + Node features  & 0.05\% & 0.4\% & 0.01\% & 0.7\% & & 43.8  & 36.0  & 45.0  & 37.4  \\
    & + \textbf{KG2Vec}  & 0.05\% & 0.4\% & 0.01\% & 0.7\% & & 44.6  & \bf 50.5  &  45.8  & \bf 51.6  \\

    \bottomrule
  \end{tabular}
  }
  \caption{Performance of OOV models on Stanford Rare Word Similarity and Card-660 datasets. Two word dictionaries are used: ConceptNet and GloVe. The overall best are underlined for each column, and the best results for each type of word dictionary are in bold. We run the BoS experiments with the default hyper-parameters from \citet{Zhao18}. Performances of other baseline models are collected from \citet{pilehvar2018}.
  }
  \label{tab:results}
\end{table*} 

Experiment results, measured by Pearson's and Spearman's correlation, on the Card-660 and Stanford rare words datasets are shown in \cref{tab:results}. The Wikipedia pages and Wiktionary definitions used in the following experiments are snapshots from Feb 16th, 2019. We compare KG2Vec to other embedding imputation models, including Mimick  \cite{PinterGE17}, Definition centroid \cite{HerbelotB17}, Definition LSTM \cite{BahdanauBJGVB17}, SemLand \cite{Pilehvar17} and BoS \cite{Zhao18}. During evaluation, zero vectors are assigned to missing words and word-word similarity is computed as the inner product of the corresponding embeddings. In KG2Vec, the number of iterations $T = 3$ for GCN, and the number of nodes with pre-trained word vectors $|V^\prime| = 9000$. We test on two types of pre-trained word vectors GloVe (Common crawl, cased 300d) and ConceptNet Numberbatch (300d). KG2Vec shows competitive performance in all test cases. On Card-660 dataset KG2Vec achieves state-of-the-art results by a significant margin. When using ConceptNet embeddings, KG2Vec results in improvements of 7.7\% and 6.7\% on Pearson's and Spearman's correlation coefficients, respectively, when compared to prior state-of-the-art performance (BoS). When using GloVe embeddings, KG2Vec improves upon SemLand by 11\% and 17.8\% on Pearson's and Spearman's correlation coefficients. Considering the fact that Card-660 contains a significant amount of recent OOV words (e.g. ``Brexit''), this improvement indicates that KG2Vec's can leverage up-to-date information from knowledge bases. Additionally, this shows that GNNs can effectively cover OOV words and precisely model their semantic meanings. On Stanford Rare Word dataset, KG2Vec is comparable with other state-of-the-art models, suggesting its robustness across various test schemes. Note that the graph used in KG2Vec has a much smaller size compared with knowledge graphs used in SemLand, the WordNet, which has 155,327 words.

To fairly evaluate KG2Vec, we include a baseline model that assigns the node feature $f_v$ as the final word representations for word $w_v$ if $w_v$ is not in the pre-trained dictionary. The results are denoted as ``Node features'' in \cref{tab:results}. In all test cases, KG2Vec improves by a large margin upon this baseline. 
For example, using GloVe on the Card-660 dataset, KG2Vec's achieves a performance increase of 14.5\% and 14.2\% respectively for Pearson's and Spearman's coefficients over Node features. This observation suggests that the information aggregation by GNN is critical for embedding imputation and semantic inference. It also indicates that learning from the knowledge graph and its language information is an effective way to parse the semantic meaning of a rare word.

\section{Discussion}
\textbf{Application on Entity Relations Knowledge Base.} Many public knowledge bases consist of relational data in a tuple format: (entity1, entity2, relation), where entities can be considered as the ``nodes'' in the graph and relations define the edges. Note that there are different kinds of relations and therefore edges in the graph have different types or labels. To impute the embeddings for entities in such scenario, one can conveniently adapt KG2Vec following \citet{schlichtkrull2018modeling} by learning different  transformations for different types of edges. 

\textbf{Adaption to New Vocabularies and Information.}
Considering the fast growth of vocabularies in the current era, the ability to perform online learning and quick adaptation for embedding imputations is a desired property. One can combine KG2Vec with meta-learning, e.g., MAML in \citet{finn2017model}, such that the resulting model can quickly learn the embeddings of newly added nodes (words), or updated node features.

\section{Conclusion and Future Work}
In this paper, we introduce KG2Vec, a graph neural network based approach for embedding imputation of OOV words which makes use of grounded language information. Using publicly available information sources like Wikipedia and Wiktionary, KG2Vec can effectively impute embeddings for rare or unseen words. Experimental results show that KG2Vec achieves state-of-the-art results on the Card-660 dataset. Future research directions include a theoretical explanation of KG2Vec and applications to downstream NLP tasks.

\section*{Acknowledgments}
We would like to thank the anonymous reviewers for their valuable feedback.


\bibliography{acl2019}
\bibliographystyle{acl_natbib}

\appendix

\end{document}